# Revolutionizing Traffic Management with AI-Powered Machine Vision: A Step Toward Smart Cities

Seyed Hossein Hosseini DolatAbadi
Department of Computer Engineering
University of Isfahan
Isfahan, Iran
S.H.Hosseini@mehr.ui.ac.ir

Sayyed Mohammad Hossein Hashemi
Department of Computer Engineering
University of Isfahan
Isfahan, Iran
mhtrxz@gmail.com

Mohammad Hosseini
Department of Computer Engineering
Ilam University
Ilam, Iran
40013119812@ilam.ac.ir

Moein-Aldin AliHosseini
Department of Software Engineering
University of Isfahan
Isfahan, Iran
Moeinaldin2022@gmail.com

*Abstract*— The rapid urbanization of cities and increasing vehicular congestion have posed significant challenges to traffic management and safety. This study explores the transformative potential of artificial intelligence (AI) and machine vision technologies in revolutionizing traffic systems. By leveraging advanced surveillance cameras and deep learning algorithms, this research proposes a system for real-time detection of vehicles, traffic anomalies, and driver behaviors. The system integrates geospatial and weather data to adapt dynamically to environmental conditions, ensuring robust performance in diverse scenarios. Using YOLOv8 and YOLOv11 models, the study achieves high accuracy in vehicle detection and anomaly recognition, optimizing traffic flow and enhancing road safety. These findings contribute to the development of intelligent traffic management solutions and align with the vision of creating smart cities with sustainable and efficient urban infrastructure.

## I. Introduction

In today's world, the rapid urbanization and population growth in cities have led to overcrowded highways and an overwhelming number of vehicles on the roads. As urban centers expand, the infrastructure often struggles to keep pace with the increasing demand[1]. Congested traffic conditions have become a daily reality for millions of commuters, resulting in wasted time, increased fuel consumption, and elevated stress levels[2]. Furthermore, the fast-paced lifestyle of modern society often leads to a disregard for traffic regulations. Drivers rushing to meet deadlines or distracted by electronic devices contribute to hazardous situations on the road [3]. These challenges underline the urgent need for innovative solutions that can effectively address traffic congestion and enhance safety.

The advent of Artificial Intelligence (AI) and Machine Learning (ML) has opened up unprecedented opportunities to tackle these challenges[4]. AI, with its capacity for analyzing massive datasets, and ML, which enables systems to learn and adapt over time, have become powerful tools in various fields, including traffic management. Machine vision, a subset of AI, utilizes advanced algorithms and cutting-edge cameras to process and interpret visual data. This technology has demonstrated remarkable capabilities in identifying objects such as vehicles and pedestrians, monitoring traffic flow, and even detecting dangerous driver behaviors like sudden lane changes or tailgating. The ability to process vast amounts of real-time data with accuracy and speed has made AI-driven machine vision an indispensable asset for modern traffic systems, with the potential to revolutionize how cities manage their roadways[5].

This study proposes a comprehensive approach to detecting and managing road overcrowding through the use of artificial intelligence (AI) and real-time data analytics. The system is designed to monitor the total number of vehicles on a given street, compare the count against a predefined vehicle threshold for that roadway, and perform mathematical analyses to determine overcrowding levels. Upon detecting an overcrowded street, the system provides immediate alerts to drivers on highways via large-screen monitors. This dynamic warning system empowers drivers to make informed route decisions, reducing congestion and optimizing traffic flow.

In the following sections, we will explore the related work that underpins this study, the approach adopted to develop the system, the evaluation of its performance, and finally, the conclusions and implications of this research.

## II. RELATED WORK

In this section, we will review a range of research focused on advancing traffic management through innovative technologies, particularly in the field of computer vision and artificial intelligence. The studies discussed explore solutions for real-time vehicle detection, traffic event recognition, and intelligent traffic management. Key areas include improvements in vehicle counting, accident detection, and traffic pattern prediction, as well as the integration of scalable computing architectures like cloud and edge computing. We will also examine advancements in road infrastructure monitoring, such as the automatic detection of surface issues and violations, as well as the development of robust systems for recognizing traffic signs and managing lane-based traffic data. Each study offers insights into overcoming challenges in modern traffic systems, with a focus on enhancing safety, efficiency, and response times in urban environments.

Bhasin et al. emphasize on how advanced computer vision technologies, powered by AI, can offer more efficient, accurate, and responsive solutions for real-time traffic conditions, such as precise vehicle counting, detection of traffic events like accidents and violations, traffic pattern prediction, and intelligent traffic light management. Leveraging scalable cloud and edge computing makes real-time traffic data analysis feasible, promising to enhance urban traffic management and transportation system efficiency and safety in the future[6].

Rahman et al. address the challenges of urban traffic management, particularly in densely populated areas, by proposing an intelligent detection system. The method utilizes the YOLOv5 model to identify vehicles in highly congested images. The authors trained four distinct versions of YOLOv5 and combined their outputs using Non-Maximum Suppression (NMS) to enhance detection accuracy[7]. Their model was tested on a dataset containing images of vehicles from crowded streets, achieving strong performance under various lighting conditions, including day and night. This approach is designed for real-time traffic management and analysis, offering potential benefits such as reduced congestion and improved traffic management.

Hu et al. propose a comprehensive framework for estimating traffic density using surveillance cameras[8]. Their system is tailored to overcome challenges like low resolution, low frame rates, limited labeled data, and complex road conditions. The framework consists of two primary components: (1) camera calibration using the MVCalib method for precise road length estimation and (2) vehicle detection utilizing transfer learning and data fusion from multiple sources to enhance model performance. The proposed models were evaluated on real-world data from Hong Kong and Sacramento, outperforming existing methods with approximately 90% accuracy in vehicle detection and a calibration error of less than 0.2 meters. The framework demonstrated effective performance in high-traffic environments, though its accuracy diminished in extremely crowded conditions. This system provides valuable traffic insights without requiring hardware upgrades.

Rithish et al. explore using computer vision to enhance road infrastructure monitoring by detecting road surface issues and traffic violations through video analysis. The research aims to improve system accuracy and efficiency by collecting diverse road data and utilizing machine learning. It discusses the importance of road infrastructure and advanced technologies in reducing traffic accidents. The study moves beyond surface defects and sign recognition to include identifying hazardous road conditions. The researchers use a large, diverse dataset to improve model generalizability, comparing image processing and machine learning methods for solving these challenges[9].

Adewopo et al. address issues with traffic accidents and analysis, essential for intelligent transportation systems and smart cities. The study presents a framework using traffic surveillance cameras and pattern recognition to detect and respond to accidents, utilizing data from the "NHTSA Crash Report Sampling System" in the U.S. It highlights the global impact of traffic accidents, with 1.35 million fatalities annually, and emphasizes the economic, medical, and infrastructure costs, particularly in developing countries. The paper discusses advanced technologies like automated accident detection systems, which improve response times and reduce human error, although challenges like standardization and system quality remain, especially in developing countries [10].

Lim et al. propose a real-time, light-resistant method for detecting traffic signs using GPGPU technology. The approach integrates object detection, a hierarchical recognition model, and parallel processing to provide stable performance in low-light conditions, achieving an F1-score of 0.97 on a traffic sign dataset. It emphasizes the importance of Traffic Sign Recognition (TSR) for autonomous vehicles, overcoming challenges in low-light and variable conditions. By using Byte-MCT features, GPGPU-based processing, and a combination of SVM and CNN for sign recognition, this method offers high accuracy and speed under various lighting conditions [11].

TABLE I. COMPARATIVE ANALYSIS OF RELATED WORK IN TRAFFIC MANAGEMENT USING ADVANCED TECHNOLOGIES

| Study | Objective | Technologies/Methods Used | Key Features/Contributions | Limitations/Gaps Addressed |
|---|---|---|---|---|
| [6] Bhasin et al. | Efficient real-time traffic management | AI-powered computer vision, cloud and edge computing | Precise vehicle counting, accident detection, traffic pattern prediction, intelligent traffic light management | Promises scalability and responsiveness but lacks implementation details in diverse conditions |
| [7] Rahman et al. | Detect vehicles in densely populated areas for real-time traffic management | YOLOv5, Non-Maximum Suppression (NMS) | Combines multiple YOLOv5 models for improved accuracy, robust performance under various lighting conditions (day/night). | Limited testing scope, performance may vary in extremely dynamic traffic or environmental conditions. |
| [8] Hu et al. | Estimate traffic density using surveillance cameras | MVCalib for camera calibration, transfer learning, data fusion | High accuracy (90%) in vehicle detection, minimal calibration error (<0.2 m), effective even in heavy traffic. | Reduced accuracy in extremely crowded environments, relies on pre-trained models and labeled data for transfer learning. |
| [9] Rithish et al. | Enhance road infrastructure monitoring | Computer vision, machine learning, large and diverse dataset | Detects road surface issues, traffic violations, and hazardous conditions; improves model generalizability | May not address real-time processing needs, particularly in high-traffic areas |
| [10] Adewopo et al. | Detect and respond to traffic accidents | Traffic surveillance cameras, pattern recognition, NHTSA Crash Report Sampling System | Reduces human error, improves response time, highlights accident costs and risks globally | Faces challenges with standardization and varying system quality, especially in developing countries |
| [11] Lim et al. | Real-time traffic sign recognition (TSR) | GPGPU-based parallel processing, Byte-MCT for low-light candidate detection, SVM for initial classification, CNN for precise recognition | Achieves F1-score of 0.97, overcomes challenges in low-light conditions, provides high accuracy and speed | Limited to traffic sign recognition, does not integrate broader traffic management needs |
| [12] Pan et al. | Vehicle detection and counting | Background subtraction, edge detection, automated lane segmentation, adaptable windows for counting | Eliminates fixed-window counting issues; adaptable windows based on lane width for more accurate counting | Limited focus on vehicle counting, no real-time accident or traffic flow detection |
| This Paper | Real-time detection of overcrowded streets and dynamic driver alerts | AI-powered vehicle counting, predefined thresholds, real-time congestion warnings | Accurate vehicle counting; dynamically notifies drivers via highway monitors; reduces congestion | Limited testing on streets with highly irregular layouts; requires integration with city-wide monitoring systems |

Pan et al. propose a system for vehicle detection and counting. The region of interest (ROI) is extracted from traffic video using background subtraction and edge detection methods. Following detection, an automated lane segmentation method is introduced. The system replaces traditional fixed-window counting methods with adaptable windows, adjusting their width based on lane width, eliminating the effect of vehicle distance and improving counting accuracy[12].

To provide a clear comparison of these contributions, we summarize key aspects of prior work in TABLE I. This table highlights the objectives, methodologies, and limitations of each study, offering a structured view of advancements and gaps in the field.

### III. APPROACH

This research leverages machine vision technology to detect and analyze various types of vehicles and traffic events using surveillance cameras. The methodology encompasses data collection and labeling, preprocessing and augmentation, and model selection and training, culminating in the development of a robust deep learning-based traffic detection system.

#### A. Data Collection

The project began by gathering a diverse dataset of traffic images from 10 strategically selected locations in Isfahan province. These locations included high-traffic areas such as Kaaveh Boulevard, Chamran Expressway, Kharazi Expressway, and a busy taxi stand. Images were captured from multiple angles to closely replicate the perspectives of real-world surveillance cameras used in traffic monitoring.

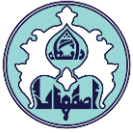 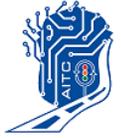

The First Biennial National Conference on the Application of Artificial Intelligence in Traffic Control

25-26 Feb. 2025, University of Isfahan, Iran

This dataset was designed to encompass a wide variety of vehicles (e.g., cars, motorcycles, trucks, buses) and environmental conditions such as different times of day, weather variations, and levels of congestion. A total of 300 images were collected, ensuring that the dataset adequately represented the diversity of real-world traffic scenarios.

*B. Data Labeling*

Each image underwent a meticulous labeling process to annotate critical features, including:

Vehicle Types: Cars, motorcycles, trucks, buses, and others were identified.

Traffic Events: Specific occurrences such as accidents, sudden stops, and traffic congestion were labeled.

Bounding Boxes: The positions and dimensions of bounding boxes for all identified objects were manually defined.

Advanced labeling tools were used to ensure the accuracy and reliability of the dataset. These detailed annotations were critical for training the deep learning models to effectively recognize and classify objects in complex traffic environments.

*C. Preprocessing and Augmentation*

To enhance the dataset and improve the model's generalization ability, preprocessing and augmentation techniques were applied. During preprocessing, images were auto-oriented and resized to a uniform resolution of 640x640 pixels, which is compatible with the requirements of YOLO models.

Augmentation strategies created ten variations of each image to increase dataset diversity. The dataset was then divided into three subsets:

- Training: 2400 images for model training.
- Validation: 300 images for hyper parameter tuning and performance monitoring.
- Testing: 300 images for evaluating the final model's accuracy and robustness.

*D. Model Selection and Training*

Two deep learning models, YOLOv11 and YOLOv8, were selected for evaluation based on their established effectiveness and popularity in real-time object detection tasks, particularly for scenarios requiring fast and accurate identification of objects in dynamic environments. Both models were carefully configured to optimize their performance, with key parameters meticulously tailored to align with the requirements of the project.

The AdamW optimizer was employed to fine-tune model weights, offering automated adjustments to both the learning rate and momentum, which are critical for ensuring efficient and stable training. The input image size was standardized to 640x640 pixels, a dimension that balances computational efficiency with sufficient detail for object detection. The models underwent 100 epochs of training, a process that allowed the networks to iteratively improve their ability to generalize patterns from the training data. Additionally, Automatic Mixed Precision (AMP) was utilized to optimize GPU memory usage, significantly reducing resource consumption and accelerating training without compromising accuracy.

To enhance stability during the initial phases of training, certain model layers were frozen, particularly those responsible for extracting foundational features, allowing the remaining parameters to focus on higher-level refinements. Among the two models, YOLOv8 stood out as the more advanced, featuring a refined architecture that prioritized parameter efficiency and improved detection capabilities. Its design enabled it to handle complex scenarios with greater precision and adaptability. The entire training process was meticulously monitored and recorded using TensorBoard, a powerful visualization tool that provided detailed insights into model performance, including loss curves, accuracy metrics, and hyper parameter trends. These records were invaluable in diagnosing potential issues, refining configurations, and ensuring the models reached their optimal performance potential.

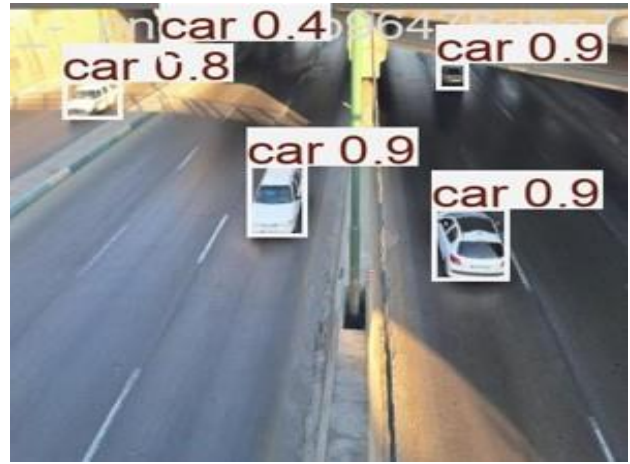

TABLE I. FIGURE 1. AN EXAMPLE OF OBJECT DETECTION

*E. Backend Analysis and Driver Notification*

The gathered data from the trained models was processed in the backend to provide an accurate assessment of traffic conditions across different districts. For every monitored street, the total vehicle count was continuously compared against a predefined threshold, which represented the

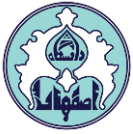
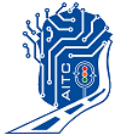

The First Biennial National Conference on the Application of Artificial Intelligence in Traffic Control

25-26 Feb. 2025, University of Isfahan, Iran

maximum allowable vehicle capacity for that specific roadway. These thresholds were determined based on road size, lane count, and historical traffic data, ensuring that the limits reflected realistic capacity constraints. The backend system then used these comparisons to calculate congestion levels dynamically and flagged streets where overcrowding was detected.

This processed information was then integrated into a sophisticated communication infrastructure designed to relay real-time updates directly to drivers. Large-screen monitors, strategically installed at key entry points to highways and major roads, displayed clear and actionable messages. These messages included warnings about specific streets that were overcrowded, estimated delay times, and alternative routes that drivers could take to avoid congestion. The system was designed with a user-centric approach, ensuring that the information was concise, easy to understand, and updated frequently to reflect changing traffic conditions.

To further enhance its effectiveness, the backend incorporated predictive analytics based on historical traffic trends and current conditions. This allowed the system to provide proactive recommendations, such as anticipating potential congestion on certain streets and preemptively advising drivers to avoid those areas. The integration of these features not only mitigated traffic in overcrowded zones but also optimized traffic distribution across the city, reducing overall delays and improving the flow of vehicles.

By bridging real-time analytics with effective driver communication, this system demonstrated a significant step forward in intelligent traffic management. It empowered drivers to make informed decisions, minimized the stress associated with traffic jams, and contributed to a more sustainable and efficient urban mobility ecosystem.

IV. EVALUATION

The proposed system was thoroughly evaluated to assess its effectiveness in detecting vehicles and traffic anomalies. The evaluation considered several factors, including overall performance metrics, class-specific accuracy, and anomaly detection capabilities.

The evaluation utilized industry-standard metrics, such as precision, recall, and mean Average Precision (mAP), to assess the models. YOLOv8 demonstrated superior performance across all metrics:

While YOLOv11 achieved slightly higher precision, YOLOv8 outperformed it in recall and mAP, making it more suitable for real-world applications where both detection accuracy and coverage are critical.

The models' ability to detect individual vehicle classes was also assessed. YOLOv8 excelled across multiple categories. In contrast, YOLOv11 struggled with certain classes, particularly buses and vans, where recall rates were significantly lower.

*A. Anomaly Detection*

One of the critical evaluation goals was to determine the system's ability to detect traffic anomalies such as accidents, sudden stops, and congestion. YOLOv8 excelled in real-time detection of these events, leveraging its faster processing speed and enhanced feature extraction capabilities. This aspect of the model has significant implications for traffic management, allowing authorities to respond to incidents more effectively.

*B. Comparative Insights*

Overall, YOLOv8 demonstrated clear advantages in terms of accuracy, adaptability, and computational efficiency. Its superior mAP scores, combined with its ability to handle diverse conditions, position it as the ideal choice for deployment in urban traffic management systems.

TABLE II. COMPARISON BETWEEN YOLOv8 AND YOLOv11 PERFORMANCE

| Model | Precision | Recall | mAP50 | mAP50-95 |
|---|---|---|---|---|
| YOLOv8 | 86.1% | %73.0 | %87.4 | %68.2 |
| YOLOv11 | %89.7 | %72.3 | %81.3 | %62.4 |

V. CONCLUSION

This research highlights the critical role of machine vision and deep learning in addressing modern traffic management challenges. By employing advanced technologies such as YOLOv8 and YOLOv11, this study developed a robust system capable of detecting vehicles and traffic events in real time with high accuracy. The approach included a comprehensive pipeline of data collection, labeling, preprocessing, and augmentation, ensuring the development of a diverse and representative dataset that reflects real-world traffic scenarios.

The results of the evaluation clearly demonstrated the superiority of YOLOv8 in terms of precision, recall, and mean Average Precision (mAP) metrics. The model's ability to adapt to diverse environmental conditions and detect various traffic anomalies, such as accidents and congestion, underscores its practicality for real-world applications. Furthermore, the use of preprocessing and augmentation techniques significantly enhanced the model's generalization capabilities, enabling it to perform effectively in complex and dynamic environments.

The findings of this research provide a strong foundation for integrating AI-driven machine vision systems into modern traffic management frameworks. Such systems can empower traffic authorities with real-time insights, improve roadway safety, and reduce congestion by enabling faster and more accurate decision-making. This approach aligns with the vision of smart cities, where intelligent technologies harmonize to optimize urban infrastructure and improve the quality of life for all road users. Future work could focus on expanding the dataset, incorporating additional environmental variables such as adverse weather conditions, and exploring further optimizations to enhance the scalability and deployment of the system.